\documentclass[journal]{IEEEtran}
\usepackage{amsmath,amsfonts}
\usepackage{algorithmic}
\usepackage{algorithm}
\usepackage{array}
\usepackage[caption=false,font=normalsize,labelfont=rm,textfont=rm]{subfig}
\usepackage{textcomp}
\usepackage{stfloats}
\usepackage{url}
\usepackage{verbatim}
\usepackage{graphicx}
\usepackage{cite}
\usepackage{multirow}
\usepackage{float}
\usepackage{orcidlink} 

\makeatletter
\newcommand{\linebreakand}{%
  \end{@IEEEauthorhalign}
  \hfill\mbox{}\par
  \mbox{}\hfill\begin{@IEEEauthorhalign}
}
\makeatother
\DeclareRobustCommand*{\IEEEauthorrefmark}[1]{%
    \raisebox{0pt}[0pt][0pt]{\textsuperscript{\footnotesize\ensuremath{#1}}}}

\begin{document}

\title{AFGI: Towards Accurate and Fast-convergent Gradient Inversion Attack in Federated Learning}
 
\author{
Can~Liu$^{\orcidlink{0009-0005-9429-6512}}$,
Jin~Wang$^{\orcidlink{0000-0003-0766-9906},}$\IEEEauthorrefmark{*},~\IEEEmembership{Member,~IEEE, }
Yipeng~Zhou$^{\orcidlink{0000-0003-1533-0865}}$,~\IEEEmembership{Member,~IEEE} 
 Yachao~Yuan$^{\orcidlink{0000-0001-7498-002X}}$,
 Quanzheng~Sheng$^{\orcidlink{0000-0002-3326-4147}}$,~\IEEEmembership{Member,~IEEE} 
and~Kejie~Lu$^{\orcidlink{0000-0002-6315-2031}}$,~\IEEEmembership{Senior Member,~IEEE}
\IEEEcompsocitemizethanks{
\IEEEcompsocthanksitem * Corresponding author: Jin Wang, Email: wjin1985@suda.edu.cn
\IEEEcompsocthanksitem Can~Liu  is with the Department of
Computer Science and Technology, Soochow University, Suzhou, China.
\IEEEcompsocthanksitem Jin Wang and  Yachao Yuan are with the School of Future Science and Engineering, Soochow University, Suzhou, China.
\IEEEcompsocthanksitem Yipeng~Zhou and Quanzheng~Sheng are School of Computing, Macquarie University, Sydney, Australia.
\IEEEcompsocthanksitem  Kejie Lu is with the Department of
Computer Science and Engineering, University of Puerto Rico at Mayag\"{u}ez, Puerto Rico,
USA.
 }}

\markboth{Journal of \LaTeX\ Class Files,~Vol.~14, No.~8, July~2024}%
{Shell \MakeLowercase{\textit{et al.}}: A Sample Article Using IEEEtran.cls for IEEE Journals}


\maketitle

\begin{abstract}
Federated learning (FL) empowers privacy-preservation in model training by only exposing users' model gradients. Yet, FL users are susceptible to gradient inversion attacks (GIAs) which can reconstruct ground-truth training data such as images based on model gradients. However, reconstructing high-resolution images by existing GIAs  faces two challenges: inferior accuracy and slow-convergence, especially when duplicating labels exist in the training batch. To address these challenges, we present an $\textbf{A}$ccurate and $\textbf{F}$ast-convergent $\textbf{G}$radient $\textbf{I}$nversion attack algorithm, called $\textbf{AFGI}$, with two components:  Label Recovery Block ($\textbf{LRB}$) which can accurately restore duplicating labels of private images based on exposed gradients; $\textbf{VME}$ Regularization Term, which includes the total \textbf{v}ariance of reconstructed images, the discrepancy between three-channel \textbf{m}eans and \textbf{e}dges, between values from exposed gradients and reconstructed images, respectively. The $\textbf{AFGI}$ can be regarded as a white-box attack strategy to reconstruct images by leveraging labels recovered by $\textbf{LRB}$. In particular, $\textbf{AFGI}$ is efficient that accurately reconstruct ground-truth images when users' training batch size is up to 48. Our experimental results manifest that $\textbf{AFGI}$ can diminish 85\% time costs while achieving superb inversion quality in the ImageNet dataset. At last, our study unveils the shortcomings of FL in privacy-preservation, prompting the development of more advanced countermeasure strategies. 
\end{abstract}


\section{Introduction}
\label{sec:intro}

\IEEEPARstart{T}{he} Federated Learning (FL) framework was originally proposed by Google~\cite{konevcny2016federated} for protecting user privacy. In FL,  users only expose updated model gradients to the parameter server (PS) for completing model training. Meanwhile, original training data is privately retained by users  on local devices to preserve privacy. In FL, the model is trained in an iterative manner, coordinated by the PS. Specifically, the PS distributes the latest global model to a set of selected  users at the beginning of each global iteration. Then, selected users train the model  with their local data for a few round of local iterations, and then return model gradients or weights back to the PS. The PS aggregates these gradients to update the global model. The above process will be repeated until 
the global model converges~\cite{FedAVG, FedSGD,  9925088, 10125048, 10032626, zhu2019deep, zhao2020idlg, geiping2020inverting, yin2021see, dong2021deep,  hatamizadeh2023gradient, zhu2020r, chen2021understanding}.  


However, recent studies have unveiled that it is unrealistic to protect data privacy by FL alone if the PS is honest-but-curious. 
According to~\cite{zhu2019deep, zhao2020idlg, geiping2020inverting, yin2021see, dong2021deep,  hatamizadeh2023gradient, zhu2020r, chen2021understanding, yue2023gradient, liu2024raf}\footnote{the paper~\cite{liu2024raf} is the previous version of the paper we published on the arXiv. This article has been modified and improved on this basis.}, gradients can leak local data privacy such that ground-truth data can be reconstructed if the honest-but-curious PS launches gradient inversion attacks (GIAs) based on collected user gradients. 
To date, there are mainly two types of GIAs: iteration-based GIAs~\cite{zhu2019deep, zhao2020idlg, geiping2020inverting, yin2021see, dong2021deep,  hatamizadeh2023gradient, yue2023gradient} and recursion-based GIAs~\cite{ zhu2020r, chen2021understanding}. 
Iteration-based GIAs reconstruct images by minimizing the distance between gradients of the ground-truth image, denoted by $\nabla W$ generated from the image $x$, and gradients of the reconstructed image, denoted by $\nabla W'$ generated by the reconstructed image $\hat{x}$~\cite{zhu2019deep, zhao2020idlg, geiping2020inverting, yin2021see, dong2021deep,  hatamizadeh2023gradient, yue2023gradient}. 
In essence, 
iteration-based GIAs' strategies employ various loss functions and auxiliary regularization terms to iteratively minimize the difference between $\nabla W$ and $\nabla W'$, continuously updating $\hat x$ through model backpropagation~\cite{zhu2019deep, zhao2020idlg, geiping2020inverting, yin2021see, dong2021deep,  hatamizadeh2023gradient, yue2023gradient}. 
In particular, the ROG strategy~\cite{yue2023gradient}  achieved the state-of-the-art results on the ImageNet dataset~\cite{deng2009imagenet}. ROG utilizes the training outcomes of the encoder, decoder, and pre-trained network as regularization. By narrowing the search range for image reconstruction, it can effectively reconstruct the ground-truth data given that FL trains LeNet, VGG-7, or ResNet-18 models~\cite{lecun1998gradient,simonyan2014very,he2016deep}.
Recursion-based GIAs, on the other hand, exploit the relationship between input data, model parameters, and gradients~\cite{zhu2020r, chen2021understanding} for reconstructing private data. Yet, these strategies are more focused on the trivial case that the batch size of data for local training is 1. Due to this  limitation, most GIAs strategies are iteration-based methods because these strategies are applicable in more generic cases. 

Nonetheless, reconstructing ground-truth data such as images is non-trivial. There are at least two reasons hindering its practical implementation. First, ground-truth images $x$ and their label $y$ information are invisible in FL. As a result, GIAs algorithms when reconstructing data slowly converge and probably diverge due to the lack of accurate critical information, \emph{e.g.}, labels. For example, it takes 8.5 hours to reconstruct a single ImageNet image under the context that the batch size is 1 by running the strategy in~\cite{geiping2020inverting} with a single NVIDIA V100 GPU.
Considering that a typical FL user may own hundreds of images~\cite{FedAVG,FedSGD}, implementing GIAs for reconstructing so many images is prohibitive.  
Second, complicated FL contexts can substantially deteriorate attack accuracy. As reported in \cite{geiping2020inverting,yin2021see}, attack accuracy is inferior when the training batch size is much grater than 1 or duplicating labels exist in the batch. 

To improve the practicability of GIAs, we propose an $\textbf{A}$ccurate and $\textbf{F}$ast-convergent iteration-based $\textbf{G}$radient  $\textbf{I}$nversion attack algorithm, called $\textbf{AFGI}$, with the $ \textbf{LRB}$ ($\textbf{L}$abel $\textbf{R}$ecovery $\textbf{B}$lock) component and the $\textbf{VME}$ (total $\textbf{V}$ariance, three-channel $\textbf{M}$ean and $\textbf{E}$dge detection regularization terms). $\textbf{LRB}$ is workable under complicated FL training tasks, \emph{i.e.}, a training batch contains multiple samples with  duplicated labels. 
Specifically, inspired by \cite{yin2021see}, we exploit the column-wise sum of the last fully connected (FC) layer gradients to serve as the input to the $\textbf{LRB}$, which can enhance the discrimination between repeated and non-repeated labels in the output probability matrix, and thereby facilitate the identification of repeated labels. Then, $\textbf{VME}$ reconstructs ground-truth data $x$ based on exposed gradients and labels $y$ produced by $\textbf{LRB}$ with a much faster convergence rate due to the following two reasons. First, $\textbf{LRB}$ can provide more accurate label information $\hat y$, which can effectively avoid learning divergence when reconstructing data. Second, we introduce the three-channel mean of an image to correct the color of $\hat x$ in each channel which is more efficient than existing works using the total mean of an image. Then, we introduce the canny edge detection, which is widely used in image subject recognition~\cite{canny1986computational}, as a regularization term in $\textbf{VME}$ such that we can more accurately recover subject positions in data reconstruction with fewer iterations.

\begin{table*}[t]
\caption{Various iteration-based GIAs within horizontal federated learning frameworks}
\centering
\begin{tabular}{c|c|c|c|c|c|c|c|c}
\hline
			\multirow{2}{*}{ Methods} &{GI} &{Image}& {Extra Terms}&{Loss} & {Image}&{Maximum}&{Label } &{Label}\\ 
                   & Types&{ Initialization} &Number& {Function}&{Resolution}&{Batch size} &{Restore}&{Assume} \\ \hline
{DLG}~\cite{zhu2019deep}  &  { Iter.} & {random}   & {0} &{$\ell_2$}& {64$\times$64}& { 8  }   &{}  &{-}   \\

                    {iDLG} ~\cite{zhao2020idlg}  &      {Iter.}         &    {random}  & {0 }& {$\ell_2$}&{64$\times$64}& {8 }&{\checkmark}&{No-repeat}\\

                      GGI  \cite{geiping2020inverting}  &            {Iter.  }          &  {  random}  & {1}& { cosine}& {224$\times$224}& 8 &{}  &{Known} \\

                     CPL \cite{wei2020framework}  &       {   Iter. }             &     {red or green }& {1} & {$\ell_2$}& {128$\times$128}&  {8}           &{}& {-} \\

                      {GradInversion} \cite{yin2021see} &   {Iter.     }              & {random}  &   {6  }&  {$\ell_2$} &  {224$\times$224} &  {48 } &{\checkmark}&{No-repeat}\\
                 HGI \cite{hatamizadeh2023gradient} &  {   Iter. }             &   {random  }&   {3}&  {$\ell_2$}&  {-} &   {8   }   &{}  &   {-} \\
            
                    ROG \cite{yue2023gradient}&  {   Iter. }             &   {Image}&   {1}&  {$\ell_2$}&  {128$\times$128} &   {16  }   &{}  &   {-} \\

                     $\textbf{AFGI}$                       (Ours)    & {  $\textbf{Iter. }$      }       & { $\textbf{gray}$ } & { $\textbf{3}$ }& {$\textbf{cosine}$}& {  $\textbf{224$\times$224}$} &  {$\textbf{48}$}  & {{\checkmark} }&   {{$\textbf{None}$} }  \\
             	 \hline
\end{tabular}
\label{tab:RL}
\end{table*}

Our experimental results unequivocally demonstrate that $\textbf{LRB}$ significantly with up to 20\% improvement of label recovery accuracy. $\textbf{VME}$ can not only accurately reconstruct ground-truth data but also considerably reduce the time cost. In particular, $\textbf{AFGI}$ saves 85\% time costs compared to the baseline strategy~\cite{geiping2020inverting}. $\textbf{AFGI}$ excels on reconstructing the widely used ImageNet dataset~\cite{deng2009imagenet}, and faithfully restore individual images at a resolution of 224$\times$224 pixels. $\textbf{AFGI}$ attack is still valid even when the batch size is up to 48 images.

Our main contributions are as follows:
\begin{itemize}

\item We use part of the training model to build a new network $\textbf{LRB}$ for label recovery. The improved label recovery accuracy with $\textbf{LRB}$ in the pre-trained ResNet-50 model, without the strong assumption of no-repeated labels, surpasses previous GIAs strategies. More accurate labels help $\textbf{AFGI}$ achieve faster convergence in the iteration process.

\item We introduce two novel regularization terms in $\textbf{VME}$ to accelerate model convergence and reduce time costs. The three-channel mean helps correct the color of $\hat x$ and and the edge detection regularization term helps align the position of subjects in $\hat x$ with those in the ground truth.

\item The reconstructed images of $\textbf{AFGI}$ outperform state-of-the-art GIAs strategies. We compare three metrics in our experiments, and $\textbf{AFGI}$ performs better in most metrics.
\end{itemize} 

In the following sections, we briefly discuss recent works relevant to $\textbf{AFGI}$ in Section~\ref{RW}. The detailed $\textbf{AFGI}$ process is elaborated in Section~\ref{sec:M}. In Section~\ref{sec:exp}, we present the results of numerous experiments. Finally, the conclusion and future directions are discussed in Section~\ref{CON}.

\section{Related Work}
\label{RW}
In this section, we first briefly introduce the classification of federated learning in Section \ref{HFL}. Then, we comprehensively survey existing gradient inversion attack strategies in Section~\ref{GI} and introduce the canny edge detection in the Section~\ref{CE}.

\subsection{Federated Learning}
\label{HFL}
FL can be classified into three main types based on data and feature distribution modes: horizontal federated learning (HFL), vertical federated learning (VFL), and federated transfer learning (FTL)~\cite{yang2019federated,zhang2022survey}. HFL, which has been extensively investigated, involves users sharing identical data features but exhibiting different data distributions~\cite{nguyen2021federated}. Conversely, VFL applies when FL users share the same data distribution but own different features, with notable applications in healthcare and finance~\cite{yang2019parallel,chen2020vafl}. FTL addresses the construction of the FL framework when both data characteristics and distribution differ~\cite{chen2021fedhealth,he2020group}. Among these types, HFL is the most studied one.

\subsection{Gradient Inversion Attacks (GIAs)}
\label{GI}
Recent studies on GIAs have primarily focused on iteration-based methods in  HFL. Zhu et al.~\cite{zhu2019deep} propose Deep Leakage from Gradients (DLG) to reconstruct ground-truth data $x$ on small datasets using the ResNet-56 model by minimizing the $\ell_2$ distance between reconstructed gradients $\nabla W'$ and ground-truth gradients $\nabla W$. However, DLG initializes with random dummy data $\hat{x}$ 
and labels, $\hat{y}$, 
which can lead to convergence failures and poor reconstructed results. To address this issue, iDLG~\cite{zhao2020idlg} utilizes the negative cross-entropy value of $\nabla W$ to obtain the high accuracy reconstructe labels $\hat y$. Geiping et al.~\cite{geiping2020inverting} propose GGI, which uses cosine similarity as the cost function and adds total variation regularization to accelerate convergence. GGI assumes that labels can be obtained via cross-entropy, which requires 192,000 (192K) iterations to reconstruct high-quality and stable images. User Privacy Leakage (CPL)~\cite{wei2020framework} effectively minimizes the $\ell_2$ distance between $\nabla W'$ and $\nabla W$ by adding label-based regularization to ensure model convergence, particularly for a batch size of 1. GradInversion~\cite{yin2021see} introduces five regularization terms to accurately position objects in $\hat x$. Building on GradInversion, NVIDIA developed the Divide-and-Conquer Inversion (DCI)~\cite{dong2021deep} algorithm, incorporating new regularization terms under Generative Adversarial Networks (GANs), and then applied these techniques to achieve excellent results with X-ray images~\cite{hatamizadeh2023gradient}. Reconstructed image data from Obfuscated Gradient (ROG)~\cite{yue2023gradient} uses an encoder and decoder to accelerate model convergence under shallow training models. Although these strategies achieve visualized reconstructed images in single-label datasets, they incur heavy time costs and rely on the unreliable assumption of batch-norm regularization~\cite{yin2021see,dong2021deep,hatamizadeh2023gradient, xu2022agic}. These strategies use the negative cross-entropy value to determine labels in GIAs. However, cross-entropy is ineffective for the acquisition of images' multi-label information.
For a succinct overview, a comparative analysis of these strategies is presented in Table~\ref{tab:RL}. Images restored by GGI with a batch size of 100 pose difficulty in identification. If the 8th column is empty, it means that the paper does not propose a new label acquisition strategy. The symbol '-' denotes instances where no specific description is provided in the corresponding paper. 

In addition to iteration-based strategies, recursive-based methods such as R-GAP~\cite{zhu2020r} reconstruct inputs of each model layer from the last layer by solving linear equations to obtain the optimal solution with the minimal error. COPA~\cite{chen2021understanding} provides a deeper understanding of the objective function and underscores the limitations of various models. However, both R-GAP and COPA concentrate on image restoration in single-label datasets with the batch size of 1.

\begin{figure*}[t]
\centering
\includegraphics[width=1.7\columnwidth]{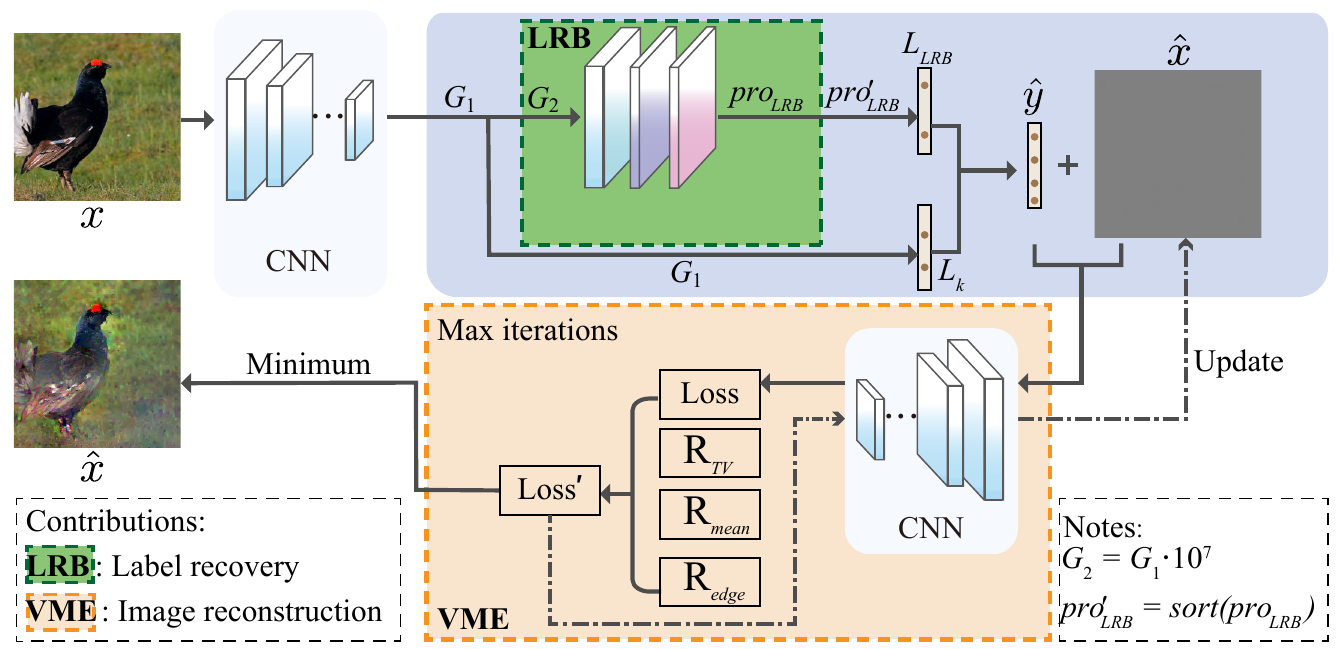} 
\caption{The workflow of $\textbf{AFGI}$. The initialization of $\hat{x}$ is a gray image and $\hat{y}$ is derived from two sources as $G_1$ ($L_k$) and the output of $\textbf{LRB}$ ($L_{LRB}$). The cosine similarity loss function with three regularization terms to compute the loss and gradients values. Finally, the $\hat{x}$ with the minimum loss value is closed to the ground-truth image.}
\label{fig:fig2}
\end{figure*}

\subsection{Canny Edge Detection}
\label{CE}
In the context of image classification, it is crucial to detect all subjects present in an image~\cite{chen2019learning, yun2021re, gao2021learning}. Various edge detection algorithms have been proposed for this purpose~\cite{canny1986computational, shrivakshan2012comparison, li2022improved}. Among these, the canny edge detection method stands out as a standard approach and remains widely used in computer vision research. Notably, it is recognized for its ability to handle gradient directions, incorporate non-maximum suppression, and utilize two thresholds to define image subject edges. Consequently, the canny edge detection method can assist GIAs in identifying subject edges, thereby potentially accelerating GIAs model convergence.


\section{AFGI Algorithm Design}
\label{sec:M}

In this section, we discuss the attack model and the workflow of $\textbf{AFGI}$. Furthermore, we elaborate the label recovery and GIAs strategies in $\textbf{AFGI}$, namely $\textbf{LRB}$ and $\textbf{VME}$.

\subsection{Attack Model}

In view of the prevalence of HFL~\cite{zhu2019deep,zhao2020idlg,geiping2020inverting,yin2021see}, our study focuses on GIAs within the HFL context. We assume that data are attacked in white-box scenarios~\cite{9904297, 10375774}. This attack context comprises an honest-but-curious PS and $N$ users $u_i$, where $i \in \{0, 1, 2, ..., N-1\}$. The PS operates according to FL framework principles but may store data during model training. It has knowledge of the model architecture, parameters, and users' uploaded gradients which is the white-box scenario~\cite{9453106, 10221704}. Consequently, the PS can construct a threat model identical to the training model and employ iteration-based GIAs to invert users' local data through model backpropagation. In more complex training tasks, \emph{i.e.}, when the training batch size is greater than 1, users' privacy is compromised even if only one training image is reconstructed. The import symbols in this paper are listed in Table~\ref{tab:sym}.

\subsection{Overview of AFGI}
\label{sec:TM}

Inspired by prior research~\cite{zhao2020idlg, geiping2020inverting, yin2021see}, we design $\textbf{AFGI}$ with two novel components: $\textbf{LRB}$, which aims to accurately restore labels of private data by using exposed gradients, and $\textbf{VME}$, which aims to reconstruct ground-truth image data efficiently.

\begin{table}[t]
\caption{Important symbols and definitions}
\centering
\renewcommand{\arraystretch}{1.2}

\begin{tabular}{c|c}
\cline{1-2}
\textbf{Symbol}                                                      & \textbf{Definition}                                           \\ \cline{1-2}
PS                                                     & the parameter server                   \\\cline{1-2}
\textbf{$x$, $y$}                                                    & the ground-truth data, label                                 \\ \cline{1-2}
\textbf{$\hat x$, $\hat y$}                                          & the reconstructed data, label                                \\ \cline{1-2}
\textbf{$\nabla W$, $\nabla W'$}                                     & the gradient of $x$ and $\hat x$                         \\ \cline{1-2}
\textbf{$\mathbf{R}$} & the regularization terms            \\ \cline{1-2}
\textbf{$\alpha$}                                            & the scaling factors      \\ \cline{1-2}
\textbf{$G_1$, $G_2$}                                                & the summation and variance of $\nabla W$        \\ \cline{1-2}
\textbf{$pro_{LRB}$}                                                 & label probability vector after $\textbf{LRB}$            \\ \cline{1-2}
\textbf{$pro'_{LRB}$}                                                & $pro_{LRB}$ in a descending order                          \\ \cline{1-2}
$K$                                                           & the training batch size                                  \\ \cline{1-2}
$N$                                                     & the number of image classes                                       \\ \hline
\textbf{$L_k$}                                                       & labels obtained from GradInversion                    \\ \hline
\textbf{$L_{LRB}$}                                                   & repeated labels obtained from $\textbf{LRB}$          \\ \hline
\textbf{$MI$}                                                        & the maximum number of iterations                              \\ \hline
\textbf{$lr$}                                                        & the learning rate                      \\ \hline
\end{tabular}

\label{tab:sym}
\end{table}

\begin{itemize}
\item $\textbf{LRB}$: A notable advantage of $\textbf{LRB}$ lies in that labels can be repeating  within one local training batch. In contrast,  prior works~\cite{geiping2020inverting, yin2021see, dong2021deep} commonly assume that no repeating labels exist in the same local training batch. In LRB, it leverages a convolutional network (by reusing the last few layers of the FL task model) to enhance the discrimination between repeating and non-repeating labels, and hence facilitate the identification of repeating labels.

\item $\textbf{VME}$: Private images are reconstructed by the PS based on recovered label information and the model backpropagation algorithm such that  $\hat{x}$ can be as close to $x$ as possible. The $\textbf{VME}$ reguarization term is involved to improve both the reconstruction accuracy and speed. 
\end{itemize}

\begin{algorithm}[h]
\caption{Label Restoration in $\textbf{LRB}$}
\label{alg: ACB}
\textbf{Input}: $G_1$: the column-wise summation of the last FC layer gradients;  $\textbf{LRB}(\cdot)$: propagation in the $LRB$; $K$: the training batch size.\\
\textbf{Parameter}: $G_2$: the result of the linear transformation of $G_1$; $\vert \cdot \vert$: the number of elements in a vector or matrix; $Con$: a large constant; $N$: the total number of classes; $sort(\cdot)$: sort $\cdot$ in decreasing order; $site(\cdot)$: obtain the index of $\cdot$; $sum(\cdot)[-1]$: summation along the column-wise; $in(\cdot)$: check if an element exists in the vector;  $L_{k}$:the reconstructed labels $\hat y$ from $G_1$ with $\vert L_{k} \vert $ as $k$; $L_{LRB}$: the $\hat y$ from $\textbf{LRB}$; $pro_{LRB}$: the probability of labels after $\textbf{LRB}$; $pro'_{LRB}$:the result of the order $pro{LRB}$; $\gamma$: the threshold of the label probability gap; $merge(\cdot)$: concatenate two vectors into one vector.\\
\textbf{Output}:  $\hat y$: the reconstructed labels.\\
\begin{algorithmic}[1] 
\STATE $G_2 = G_1\times Con$
\STATE $\vert L_{LRB} \vert = 0$
\STATE $\vert L_{k} \vert = k$
\STATE $pro_{LRB}= sum(LRB(G_2))[-1]$
\STATE $pro'_{LRB}= sort(pro_{LRB}) $
\STATE $id_{mid} =site( pro'_{LRB}) $
\IF{$k < K$}
\STATE $G_2 = G_1\times Con$
\STATE $\vert L_{k} \vert = 0$
\STATE $\vert L_{LRB} \vert = 0$
\FOR {i  = 0, ..., N-1}
\IF {$G_1[i] < 0$}
\STATE $L_{k} \leftarrow arg~ sort(G_1[i])$
\ENDIF
\ENDFOR
\STATE $\vert L_{k} \vert = k$
\STATE $pro_{LRB}= sum(LRB(G_2))[-1]$
\STATE $pro'_{LRB}= sort(pro_{LRB}) $
\STATE $id_{mid} =site( pro'_{LRB}) $
\IF{$k < K$}
\FOR {i  = 0, ..., N-1}
\IF {$id_{mid}[i]~in~L_{k} ~and $\\$ (pro'_{LRB}[id_{mid}[i]]-pro'_{LRB}[id_{mid}[i+1]]) > \gamma$}
\STATE $L_{LRB}  \leftarrow id_{mid}[i]$.
\ENDIF
\STATE $k = k+1$
\ENDFOR
\ENDIF
\ENDIF
\IF{$\vert L_{LRB} \vert < K-k$}
\STATE $L_{LRB}  \leftarrow L_k[m]$
\ENDIF
\end{algorithmic}
\end{algorithm}

\subsection{Label Recovery Block (LRB)}
\label{sec:LR}


Provided that the training batch size is $K$, where $K \in \{0, 1, ..., N\}$, our task is to recover $K$ labels contained in a user's local training batch. 
The process to recover labels  has two major steps:  extracting initial labels (denoted by a sequence $L_k$) , and obtaining duplicating labels (denoted by a sequence $L_{LRB}$) through $\textbf{LRB}$. Note that both $L_k$ and $L_{LRB}$ are sorted sequences by a descending order of probabilities of these labels contained in the training batch. 

Prior works such as GradInversion~\cite{yin2021see} can extract labels assuming that there is no repeating labels, which can be reused by our works to extract initial labels $L_k$ in the first step. Specifically, it extracts  $L_k$   from negative cross-entropy values. 
This method involves selecting the label corresponding to the minimum $K$ gradient values from the last FC layer. The GradInversion label recovery method is briefly outlined as follows:
\begin{equation}
  \label{eq:Grad}
    \hat y = argsort(\min_M~\nabla{ W_{M, N}^{(FC)}}\mathcal{L}(x, y))[:K],
\end{equation}
where $\min_{M}~\nabla_{ W{M, N}^{(FC)}}\mathcal{L}(x, y)$ represents the minimum $\nabla{W}$ value obtained from the last FC layer along the feature dimension (equivalently by rows). Here, $M$ is the number of embedded features and $N$ is the number of image classes. The function $argsort(\cdot)[:K]$ is utilized to retrieve the indices of the minimum $K$ values along a column. Finally, it outputs a sorted label sequence $L_k$ and a sorted probability sequence ${p}_{init}$. Elements in $L_k$  are sorted by a descending order of their probabilities. ${p}_{init}$ indicates the corresponding probability of each element. 
For example,  $L_{k} = (c_1, c_2, c_3)$ indicates that recovered labels include $c_1$, $c_2$ and $c_3$. If the corresponding probability of each label in the training batch is 0.9, 0.8, 0.7, respectively, it implies that  ${p}_{init} =(0.9, 0.8, 0.7)$. 

Note that there is no repeating labels in $L_{k}$.
Given the possibility of duplicating labels in the local training batch in practice, it is likely that the number of elements in  $L_k$ is less than $K$, i.e., the number of samples in the training batch. It is represented by 
$\mid L_k \mid  = k\le K$, where $\mid \cdot \mid$ denotes the number of elements in $\cdot$. 
Here, $k < K$ indicates that there exists $K-k$ repeating labels in the training batch.

The second step involves obtaining the duplicating label sequence, denoted by $L_{LRB}$, where $\mid L_{LRB} \mid = K - k$. We design $\textbf{LRB}$ as an auxiliary structure for acquiring these duplicating labels. $\textbf{LRB}$ consists of the last block, the last average pool (AvgPool) layer, and the last fully connected (FC) layer of the FL training model. 
$\textbf{LRB}$ shares the same model parameters as the corresponding layers of the model trained by FL. In previous label classification tasks~\cite{krizhevsky2017imagenet, huang2023self, yang2023language}, labels are typically identified as the largest values in the model output. Following this intuition, if label is contained in the training batch, the output of $\textbf{LRB}$ for this label should be higher than that of other labels. 

$\textbf{LRB}$ recovers duplicating labels by leveraging exposed gradients 
through the following three substeps:
\begin{itemize}
  \item \textbf {Substep 1:} The input to $\textbf{LRB}$ is the variance of exposed gradients $\nabla{W}$. Initially,  $G_1 \in \mathbb{R}^{1 \times h}$ is defined as the summation of  gradients of the last fully connected (FC) layer in a row-wise manner, where $h$ is the number of output\_channel before the last FC layer. For convenience, we multiply $G_1$ by  a large positive integer, e.g., $10^{7}$, to get $G_2$, which will not alter the numerical relationships between gradients. 
    Then, the dimension of $G_2$ is expanded from $\mathbb{R}^{1 \times h}$ to $\mathbb{R}^{1 \times h \times 1 \times 1}$ to fit the input  of $\textbf{LRB}$, which is a common operation in computer vision for adjusting the dimensions of images~\cite{liu20234d, NEURIPS2023_f188a553}. Using the expanded $G_2$ as input of $\textbf{LRB}$, a label probability sequence ${p}_{LRB} \in \mathbb{R}^{1 \times N}$ is produced, where $N$ represents the number of image classes, corresponding to the number of outputs in the last FC layer of the trained FL model.
    
 \item \textbf {Substep 2:} 
To identify labels contained in a local training batch, we sort ${p}_{LRB}$ in a descending order. 
We select a label from  $L_{LRB}$ and insert it to $L_k$ if and only if the following two conditions are met. First, the label is an element in $L_k$. The second condition depends on ${p}_{LRB}$. Supposing that the label is the $n$-th element in ${p}_{LRB}$. It will be selected if ${p}_{LRB}[n]- {p}_{LRB}[n+1]>\gamma$, where $\gamma$ is an empirically determined threshold and $[n]$ indicates the $n$-th largest element in a sequence. According to our experiments in Section~\ref{sec:exp}, $\gamma$ is set as  $0.4$. The intuition is that a label is repeating only if its probability in the training batch is much greater than other candidate labels. 

\item \textbf {Substep 3:} 
 It is possible that $\mid L_{LRB} \mid$ is still less than $K-k$ after Substep 2. We further duplicate top $K-k$ labels in $L_k$. 
 Here, top labels are selected based on their probabilities in ${p}_{init}$.
In this case, we choose the label with the highest probability value as the duplicating label. After these steps, we obtain a reconstructed label sequence $\hat y$ in increasing order, which contains $K$ labels. 
\end{itemize}

\begin{figure}[t]
  \centering
    \includegraphics[width=0.45\textwidth]{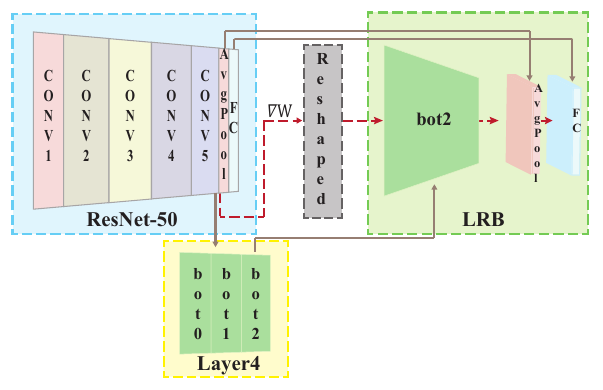}
 \caption{The layers in the LRB are based on the ResNet-50 model.}
\label{fig:ACB}
\end{figure}

The label recovery process is visually illustrated in the green box in Fig.~\ref{fig:fig2}, with detailed layers shown in Fig.~\ref{fig:ACB}, and the pseudocode provided in Algorithm~\ref{alg: ACB}.  Lines 4-8 extract $L_k$, while lines 10-19 manage the $L_{LRB}$ procedure. Ultimately, the merge of $L_k$ and $L_{LRB}$ ensures that the restoration of labels aligns with the batch size $K$. The repeated label reconstruction requires only one propagation of $\textbf{LRB}$. The computational complexity of $\textbf{LRB}$ is $\mathcal{O}((K-k) \cdot N)$.

%

\subsection{Regularization Terms $VME$}
\label{OF}
We aim to reduce time costs and enhance the quality of $\hat x$ in white-box attack scenarios. To achieve this goal, we update $\hat x$ using model backpropagation, regularized by $\textbf{VME}$. Here, recovered labels $\hat y$ and initialized gray images are employed as inputs. According to ablation experiments in GradInversion~\cite{yin2021see}, the absence of regularization terms makes bad results in $\hat x$. To improve the quality of $\hat x$, we introduce three regularization terms, i.e., total $\textbf{v}$ariance, three-channel $\textbf{m}$ean discrepancy, and $\textbf{e}$dge detection, into the $\textbf{AFGI}$ objective function:
\begin{equation}
  \label{eq:obj}
    \hat x = \mathop{argmin}\limits _{\hat{x}, \hat{y}}1 - cos( \nabla W', \nabla W ) + \mathbf{R}_{reg},
\end{equation}
where
\begin{equation}
\label{eq:cos}
   \cos( \nabla W', \nabla W ) =  \frac{< \nabla W, \nabla W'>}{\Vert  \nabla W\Vert  \Vert  \nabla W' \Vert},
\end{equation}
and
\begin{equation}
\label{eq:reg}
    \mathbf{R}_{reg} =  \alpha_{TV}\mathbf{R}_{TV} + \alpha_{mean}\mathbf{R}_{mean} +  \alpha_{ed}\mathbf{R}_{ed} .
\end{equation}
Here, Equation~\ref{eq:obj} employs the cosine similarity distance (denoted by $cos(\cdot)$ in Equation~\ref{eq:cos}) to measure the gap between $\nabla W'$ and $\nabla W$, following the previous work~\cite{geiping2020inverting}. The computational complexity of Equation~\ref{eq:cos} is \begin{math} \mathcal{O}(d) \end{math}, where $d$ is the dimension of the image data matrix. When Equation~\ref{eq:obj} approaches 0, $\hat x$ approximates $x$. In Section~\ref{sec:exp}, we will empirically compare different loss functions to demonstrate that $cos(\cdot)$ is a better choice as the loss function in our problem. 

The regularization term $\mathbf{R}_{reg}$ in $\textbf{VME}$ comprises three components: $\mathbf{R}_{TV}$, $\mathbf{R}_{mean}$, and $\mathbf{R}_{ed}$. $\mathbf{R}_{TV}$ penalizes the total variance of $\hat{x}$ and has been proven effective in numerous prior works~\cite{geiping2020inverting,yin2021see}.
Our novelty lies in introducing  $\mathbf{R}_{mean}$  and $\mathbf{R}_{ed}$.

\textbf{Mean regularization term} $\mathbf{R}_{mean}$: The design of  $\mathbf{R}_{mean}$ is inspired by our measurement study on publicly available image datasets. As shown in Table~\ref{tab:me}, we measure three-channel average of images in most popular image datasets, including 
ImageNet, CIFAR-10, CIFAR-100, PASCAL VOC 2012, and Microsoft Common Objects in Context (MS COCO). 
From Table~\ref{tab:me}, we can observe that the means of three-channel  of these image datasets are very close to each other. Thereby, it is reasonable to assume that the mean of a reconstructed image should not be far away from these mean values. The average mean value of these image datasets is calculating in the last row.
By averaging these mean values, we use [0.491, 0.467, 0.421] as the prior-knowledge to construct the term for regularizing the objective function. In other words, the $\ell_2$ distance between the three-channel mean of a reconstructed image $\hat x$ and [0.491, 0.467, 0.421] should be as small as possible. 

\begin{table}[t]
\caption{The most popular image datasets in computer vision}
\centering
\renewcommand{\arraystretch}{1.1}
\begin{tabular}{c|ccc}
\hline
\textbf{Dataset} & \multicolumn{3}{c}{\textbf{Mean values of three-channel}} \\ \hline
ImageNet         & \multicolumn{3}{c}{{[}0.485, 0.456, 0.406{]}}                 \\ \hline
CIFAR-10         & \multicolumn{3}{c}{{[}0.491, 0.482, 0.447{]}}                 \\ \hline
CIFAR-100        & \multicolumn{3}{c}{{[}0.507, 0.487, 0.441{]}}                 \\ \hline
PASCAL VOC 2012  & \multicolumn{3}{c}{{[}0.485, 0.456, 0.406{]}}                 \\ \hline
MS COCO          & \multicolumn{3}{c}{{[}0.485, 0.456, 0.406{]}}                 \\ \hline
\textbf{Average} & \multicolumn{3}{c}{\textbf{{[}0.491, 0.467, 0.421{]}}}        \\ \hline
\end{tabular}
\label{tab:me}
\end{table}

\textbf{Edge regularization term} $\mathbf{R}_{ed}$:
\label{subsec:ca} 
Prior works~\cite{geiping2020inverting,yin2021see} have shown that the position of a subject in  $\hat x$ may deviate from the counterpart in  $x$. To address this problem, $\mathbf{R}_{group}$ is introduced into the objective function of GradInversion~\cite{yin2021see} to calculate the average of  positions of a subject from different initial seeds, which can  alleviate the significant deviation. $\mathbf{R}_{group}$ achieves a better quality of $\hat x$ through the initialization of more random seeds. However,  using $\mathbf{R}_{group}$ can heavily increase the time cost of GIAs.

To overcome the heavy time cost drawback of group regularization, 
we introduce a superseded  regularization term called edge regularization ($\mathbf{R}_{ed}$) based on exposed gradients and the canny edge detection algorithm. Since $x$ is invisible and  we can only access $\nabla W$ for deriving the edge information, we modify the original canny edge detection algorithm accordingly, and the workflow to obtain $\mathbf{R}_{ed}$ is as follows:

\begin{itemize}

\item A dynamic threshold $fin$ related to the maximum and mean values of the gradient before the last FC layer $G$ ($max (G)$, $mean (G)$) is set as:
\begin{equation}
\label{fin}
    \\  fin = (max (G)-mean (G)) \times \beta.
\end{equation}
We utilize the average value ($mean(\cdot)$) to filter out gradients that are extremely small. The parameter $\beta$ takes  the value range for capturing gradients of all edges into account.

\item Obtain a set of $G$ matrix coordinates $ed_{fin}$ where the values of $\nabla W$ are greater than  $fin$.

\item Scale $G$ according to the dimension of $x$ to obtain  pixel positions $ed_{fin-reg}$ corresponding to $ed_{fin}$. Finally, choose the middle pixel position $ed_{reg}$ of $ed_{fin-reg}$ as the $x$ baseline point. 

\item Apply the canny edge detection algorithm on $\hat x$ during each backpropagation iteration with two thresholds of $\theta_1$ and $\theta_2$, obtaining a set of edges denoted as $ed_t$. As there is a significant gap between $\hat x$ and $x$ in the early stage of iterations, set these two thresholds to be sufficiently large to reduce the impact of non-edge pixels. Then, select the middle pixel coordinate $\hat {ed}_{reg}$ from $ed_t$ as the baseline point for $\hat x$.

\item Calculate  $\ell_2$ distance between $ed_{reg}$ and $\hat{ed}_{reg}$, penalizing significant differences
\end{itemize}

\begin{algorithm}[t]
\caption{Edge regularization term $\mathbf{R}_{ed}$}
\label{ca_reg}
\textbf{Input}: $G$: the gradient before the last FC layer; 
				$CA( \hat x_i, \theta_1, \theta_2)$: canny edge detection of image $\hat x$ with two thresholds $\theta_1$ and $\theta_2$.\\
\textbf{Parameter}: $max(\cdot)$: the maximum value of $\cdot$; 
	$mean(\cdot)$: the mean value of $\cdot$;
	 $\beta$: the threshold to select gradients from $G$; 
	 $fin$: the threshold to obtain edges from $G$; 
	 $ed_{fin-reg}$: the pixel site set where the gradient value is larger than $fin$;
	  $\vert \cdot \vert$: the number of elements in a vector or matrix; 
	  $img_{row}, img_{col}$: the ground-truth image matrix dimensions; 
	  $index(\cdot)$: the index of $\cdot$.\\
 \textbf{Output}: ${R}_{ed}$: the value of $\mathbf{R}_{ed}$.
 
\begin{algorithmic}[1]

\STATE   $fin = (max(G)-mean(G)) \times \beta$
\FOR{$i~from~0~to~\vert G\vert$}   
\IF{$G[i] > fin$}
\STATE{$ed_{fin} \leftarrow G[i]$}
\ENDIF 
\ENDFOR
\FOR{$j~from~0~to~\vert ed_{fin}\vert $}
\STATE $ed_{fin-reg}\leftarrow [\lfloor ed_{fin}[j]\times img_{row} / \vert ed_{fin}\vert \rfloor, \lfloor ed_{fin}[j]\times img_{col} / \vert ed_{fin}\vert \rfloor]$
\ENDFOR

\STATE $ed_{reg} \leftarrow ed_{fin-reg}[1/2 \times \vert ed_{fin}\vert]$
\STATE $ed_{t}\leftarrow CA(\hat x, \theta_1, \theta_2)$
\STATE$\hat{ed}_{reg} \leftarrow ed_{t}[1/2 \times \vert ed_t \vert]$
 \STATE ${R}_{ed} \leftarrow \lVert ed_{reg}-\hat{ed}_{reg} \rVert_2$
\STATE $ \textbf{return}~{R}_{ed}$
\end{algorithmic}
\end{algorithm}

\begin{algorithm}[ht]
\caption{Regularization Terms $\textbf{VME}$}
\label{TEA1}
\textbf{Input}: $\nabla W$: the ground-truth gradients; $\nabla W'$: the reconstructed image gradients; $\hat y$: the reconstructed labels; $MI$: the maximum iteration times; $\mathbf{R}_{TV}$: the total variance regularization term; $\mathbf{R}_{mean}$: the three-channel mean regularization term;  $\mathbf{R}_{ed}$: the canny edge detection regularization term; $\alpha_{TV}$: the scaling factor for $\mathbf{R}_{TV}$; $\alpha_{mean}$: the scaling factor for $\mathbf{R}_{mean}$; $\alpha_{ed}$: the scaling factor for $\mathbf{R}_{ed}$.\\
\textbf{Parameter}: $gray$: gray image; $cos(\cdot)$: the cosine similarity function; $\mathbf{R}_{reg}$: the value of three regularization terms; $argmin(\cdot)$: get the minimum value of $\cdot$; $M_{back}(\cdot, \cdot)$: the backpropagation of the model.\\
\textbf{Output}:  $\hat x$: the reconstructed image.\\
\begin{algorithmic}[1] 
\STATE {$\hat x = gray$}
\FOR {$i~from~0~to~MI-1$}
\STATE $\mathbf{R}_{reg} =  \alpha_{TV}\mathbf{R}_{TV} + \alpha_{mean}\mathbf{R}_{mean} +  \alpha_{ed}\mathbf{R}_{ed}$
\STATE $\hat x = \mathop{argmin}\limits _{\hat{x}, \hat{y}}1 - cos( \nabla W', \nabla W ) + \mathbf{R}_{reg}$
\STATE $M_{back}(\hat x, \hat y)$
\STATE update $\hat x$ 
\ENDFOR
\STATE $\textbf{return}~\hat x$
\end{algorithmic}
\end{algorithm}

The  pseudocode for the process to obtain  $\mathbf{R}_{ed}$  is presented in  Algorithm~\ref{ca_reg} and~\ref{ca_reg1}. Lines 1-6 depict the process of obtaining edges from  ground-truth gradients, lines 7-9 represent the workflow for the base-point of $x$, while lines 10-12 obtain $\hat x$ edges, and line 13 is the $\ell_2$ distance calculation process. The result of $\mathbf{R}_{ed}$ is used to penalize the gap between $x$ and $\hat x$.

\section{Experiments}
\label{sec:exp}
In this section, we first elaborate the settings of  experimental environment and hyperparameters. Next, we introduce three evaluation metrics used for the subsequent quantitative comparison of recovered $\hat{x}$. Following this, we compare the accuracy of $\hat{y}$ to demonstrate the superior precision of $\textbf{LRB}$. Finally, we present both visual and quantitative comparisons of $\hat{x}$ between $\textbf{AFGI}$ and other typical iteration-based GIA strategies.

\subsection{Experiments Details}

We utilize the pre-trained ResNet-50 model as the attack model, implementing the $\textbf{AFGI}$ strategy to invert images of size 224×224 pixels from the validation set of the ImageNet ILSVRC 2021 dataset~\cite{deng2009imagenet}. This setup aligns with that of GradInversion~\cite{yin2021see}. In the $\textbf{AFGI}$ experiment, we   conFig.d hyperparameters by setting $\alpha_{TV} = 1e-1$, $\alpha_{mean} = 1e-3$, $\alpha_{ed} = 1e-2$, and the learning rate $lr = 1e-2$ during the maximum iterations $MI = 10K$ at a batch size of 1. The parameters with $\alpha = 0.6$, $\theta_1 = 0.8$ and $\theta_2 = 0.9$  are used in $\mathbf{R}_{ed}$. Fine-tuning is then performed for additional 10K iterations with a batch size larger than 1. The $lr$ decreases with  a factor of 0.2  after every  2/7 of the total iterations 
in our experiments. The most time-saving and cost-effective setting for $\textbf{AFGI}$ is to restart only once, while GGI~\cite{geiping2020inverting} requires eight restarts. The inversion process of $\textbf{AFGI}$ is accelerated using one NVIDIA A100 GPU, coupled with the Adam optimizer.

\subsection{Evaluation Metrics}

In this paper, we employ three key metrics  for evaluating the quality of $\hat{x}$ from three perspectives, which can overcome the limitations of evaluation relying on a single metric.

\begin{itemize}
\item Peak Signal-to-Noise Ratio (PSNR $\uparrow$): This metric measures the ratio of the peak signal strength to the noise level, providing insights into the fidelity of $\hat{x}$. A higher PSNR value indicates a closer match to $x$.

\item Structural Similarity (SSIM $\uparrow$): SSIM assesses the structural similarity between $x$ and $\hat{x}$. A higher SSIM value signifies a more faithful reproduction of the structural details in $x$.

\item Learned Perceptual Image Patch Similarity (LPIPS $\downarrow$): LPIPS quantifies perceptual similarity between $x$ and $\hat{x}$ by considering learned image features. A lower LPIPS value indicates a higher perceptual similarity to $x$.

\item Time cost ($\downarrow$): The time cost is obtained by counting the GPU time consumed by the strategy. The less time cost, the faster the convergence speed of the strategy model and the higher the efficiency.

\end{itemize}

\begin{table}[t]
\caption{Label reconstruction accuracy on the ImageNet dataset validation set}
\centering
\renewcommand{\arraystretch}{1.1}
\begin{tabular}{c|c|c|c}
\hline
		{Batch size}&{iDLG}\cite{zhao2020idlg}&{GradInversion}\cite{yin2021see} & {$\textbf{LRB}$} $\textbf{(Ours)}$ \\ 
						
    \hline           
1  			& $\textbf{100\% }$  &  $\textbf{100\% }$&  $\textbf{100\% }$  \\
2  			& 78.45\%  			 & 87.15\% 			 &  $\textbf{98.90\% }$  \\
4                 & 68.05\%       		 & 80.50\% 			&  $\textbf{91.95\% }$\\
8                 & 58.11\%    			& 75.29\% 			&  $\textbf{87.01\% }$    \\ 
 16              & 53.42\%  			& 72.09\% 			&  $\textbf{83.95\% }$\\ 
  32             & 48.85\%  			& 68.85\%  			&  $\textbf{80.96\% }$\\ 
   64            & 46.07\%    			& 66.48\%  			&  $\textbf{79.24\% }$ \\ 
 
             	 \hline
\end{tabular}
\label{tab:LR}
\end{table}

\begin{table}[b]
\caption{Label reconstruction using $\textbf{LRB}$ under different $\gamma$}
\centering
\renewcommand{\arraystretch}{1.1}
\begin{tabular}{c|c|c|c}
\hline
					{Batch size}	&{0.3} & {$\textbf{0.4}$} &{0.5} \\ 
    \hline           
1  			& $\textbf{100\% }$  &    $\textbf{100\% }$  &$\textbf{100\% }$\\
2  			& 98.90\%  			 			 &  $\textbf{98.90\% }$  &98.75\% \\
4                 & 91.78\%       					&  $\textbf{91.95\% }$&90.98\% \\
8                 &  $\textbf{87.25\%}    $				& 87.01\%  & 86.69\% \\ 
 16              & 83.57\%  						&  $\textbf{83.95\% } $&83.83\% \\ 
  32             &$\textbf{ 81.18\% }	$		 			& 80.96\% & 81.07\% \\ 
   64            & 79.23\%    			 			& 79.24\% & $\textbf{79.29\% }$\\ 
 
             	 \hline
\end{tabular}
\label{tab:04}
\end{table}

\subsection{Label Restoration}
In label restoration experiments, we compare $\textbf{LRB}$ with the state-of-the-art methods, including GradInversion~\cite{yin2021see} and iDLG~\cite{zhao2020idlg}. The latter one was also employed in GGI~\cite{geiping2020inverting}. Images in each training batch are uniformly sampled, with the initial image index randomly chosen from 1 to 20K, and a uniform random number between 1 and 100 serving as the increment for the next image index. Label accuracy results are summarized in Table~\ref{tab:LR}. Notably, $\textbf{LRB}$ consistently outperforms iDLG and GradInversion across all batch sizes. Specifically, with  a batch size of 2, $\textbf{LRB}$ achieves 98.90\% accuracy, remarkably surpassing iDLG by more than 20\% and outperforming GradInversion by over 11\%.

\begin{figure}[t]
  \centering
  \subfloat[Ground-truth]{
    \includegraphics[width=0.08\textwidth]{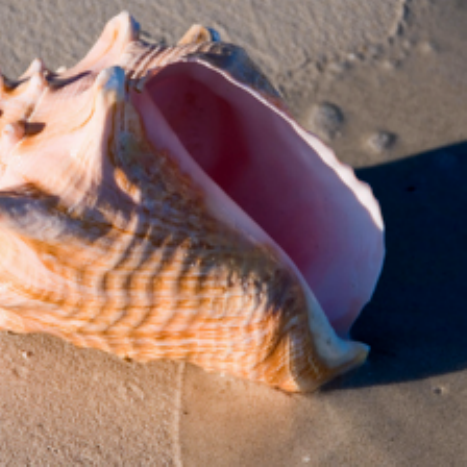}
    \label{3a}}
 \subfloat[\textbf{AFGI}]{
    \includegraphics[width=0.08\textwidth]{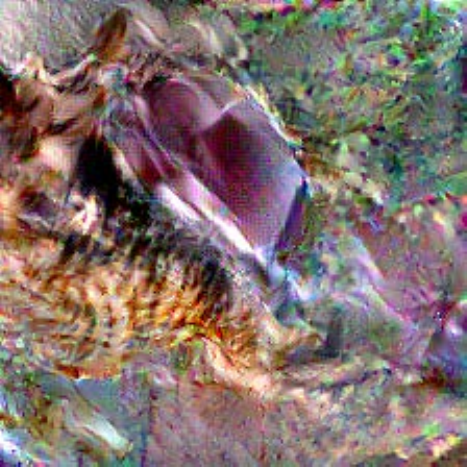}
    \label{fig:3b}
}
  \subfloat[DLG]{
    \includegraphics[width=0.08\textwidth]{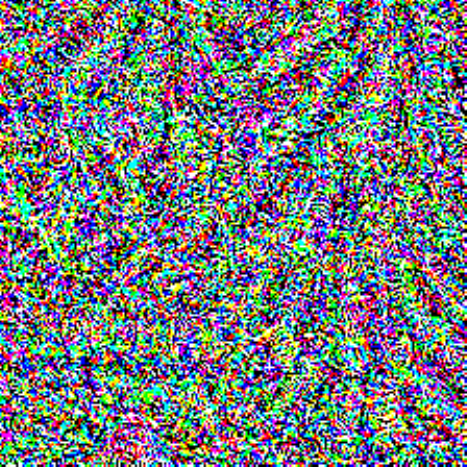}
    \label{fig:3c}
}
  \subfloat[GGI]{
    \includegraphics[width=0.08\textwidth]{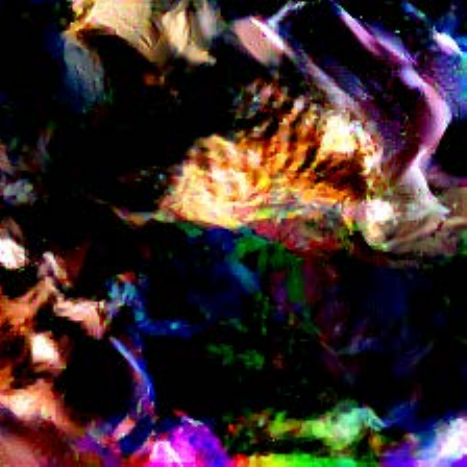}
    \label{fig:3d}
}
  \subfloat[ROG]{
    \includegraphics[width=0.08\textwidth]{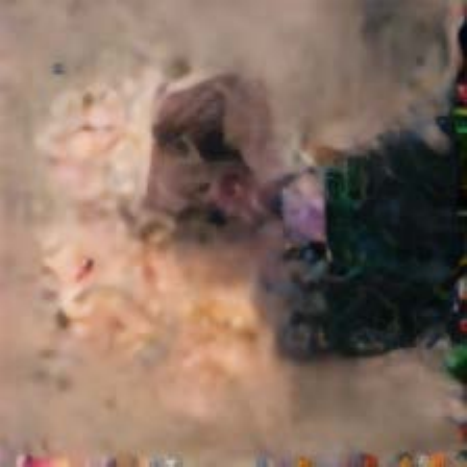}
    \label{fig:3e}
}

 \subfloat[Ground-truth]{
    \includegraphics[width=0.08\textwidth]{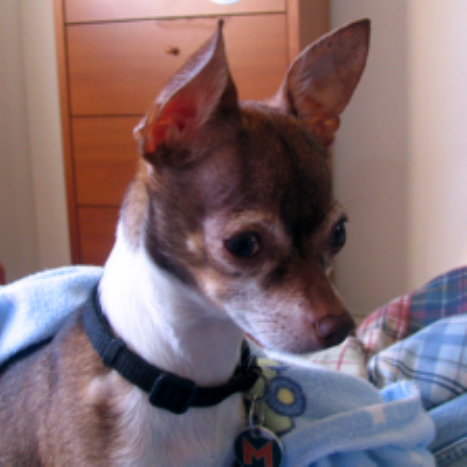}
    \label{fig:3f}
}
 \subfloat[\textbf{AFGI}]{
    \includegraphics[width=0.08\textwidth]{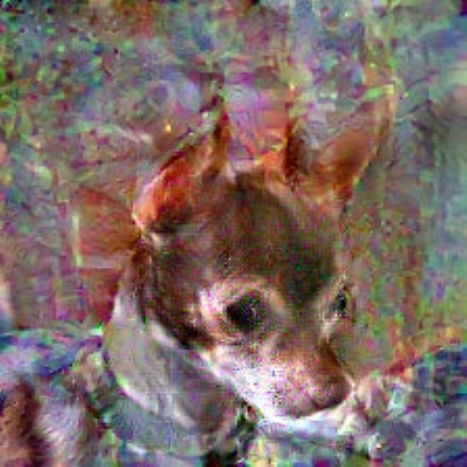}
    \label{fig:3g}
}
 \subfloat[DLG]{
    \includegraphics[width=0.08\textwidth]{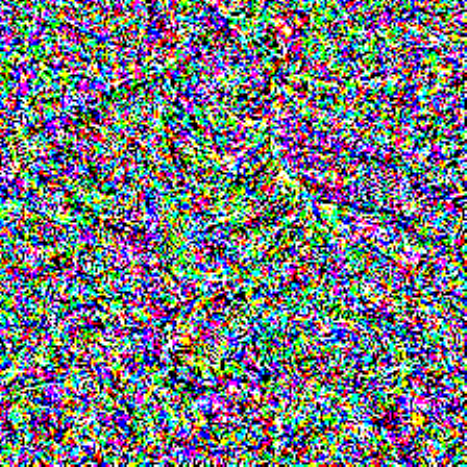}
    \label{fig:3h}
}
 \subfloat[GGI]{
    \includegraphics[width=0.08\textwidth]{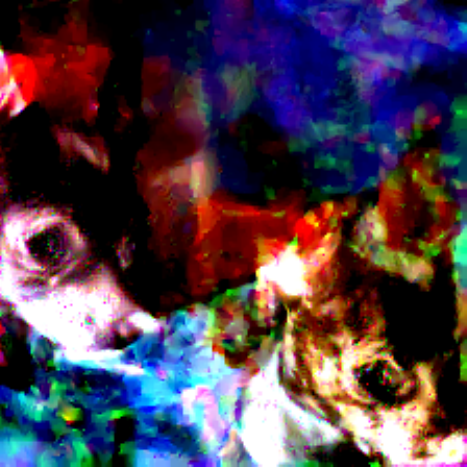}
    \label{fig:3i}
}
 \subfloat[ROG]{
    \includegraphics[width=0.08\textwidth]{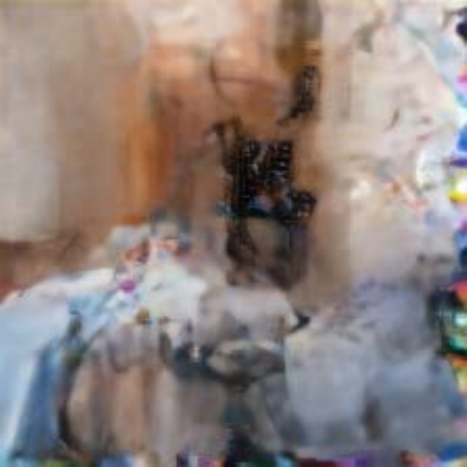}
    \label{fig:3j}}
  \caption{Comparing the results of $\hat x$ under the batch size of 1. Figures (a) and (f) depict ground-truth images $x$. Figures (b), (c), (d), (e), (g), (h), (i), and (j) showcase $\hat x$ under different GIAs strategies.}
  \label{fig:fig 3}
\end{figure}

\begin{table}[t]
\caption{The metric values of Fig.~\ref{fig:fig 3}}
\centering
\renewcommand{\arraystretch}{1.1}
\begin{tabular}{c|c|c|c|c}
\cline{1-5}
 &(b) & (c) &(d) &(e)\\ \hline
PSNR $\uparrow$  &17.47  &3.55 & 10.07&$\textbf{ 19.26}$\\
SSIM  $\uparrow$ & \textbf{0.0572} & 0.0156 &  0.0137& 0.0417 \\
LPIPS $\downarrow$ &  $\textbf{0.4909}$ & 1.4242 &  0.6788&0.6044\\\hline
          & (g) &(h) &(i) & (j)\\ \hline
PSNR $\uparrow$   &15.87   &  3.56& 11.60 &  $\textbf{17.87}$\\
SSIM $\uparrow$ &$\textbf{0.1183}$  &0.0143 &  0.0378& 0.0485\\
LPIPS $\downarrow$ & $\textbf{0.5762}$  & 1.4843 & 0.6813&  0.5970\\
\hline
\end{tabular}
\label{tab:tab3}
\end{table}

We then compare $\textbf{LRB}$ label accuracy under various $\gamma$ values in Table~\ref{tab:04}. The $\gamma = 0.4$ demonstrates superior performance across different batch sizes in terms of label recovery accuracy. Consequently, we adopted $\gamma = 0.4$ in the final scheme design.

\begin{figure}[t]
  \centering
  \subfloat[Ground-truth]{
    \includegraphics[width=0.4\textwidth]{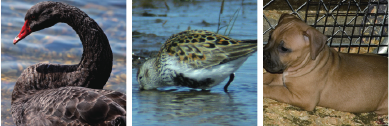}
    \label{fig:BS=1-a}
}

  \subfloat[\textbf{AFGI}]{
    \includegraphics[width=0.4\textwidth]{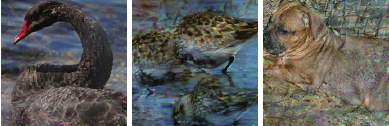}
    \label{fig:BS=1-b}
}
  \caption{The results of $\hat x$ with a batch size of 1. We reconstruct the 5000-th, 7000-th, and 9000-th images of the ImageNet validation set in $\textbf{AFGI}$.}
  \label{fig:1000th}
\end{figure}

\subsection{AFGI Reconstruct Results}
In this subsection, we present the reconstructed results under different batch sizes. When the batch size is 1, we examine the ablation experiments to test the effect of different regularization terms. Next, we compare the time costs of different GIA strategies to provide a comparison of different initialization images and loss functions. Finally, we apply  $\textbf{AFGI}$ to obtain $\hat{x}$ when the batch size is greater than 1.

\subsubsection{Reconstructed Results as Batch Size = 1}

As DLG~\cite{zhu2019deep} is a pioneering work in iteration-based GIAs, GGI~\cite{geiping2020inverting} utilizes cosine similarity as the loss function, which is the same as $\textbf{AFGI}$. Additionally, ROG~\cite{yue2023gradient} achieves state-of-the-art GIA results on the ImageNet dataset. Consequently, we perform a comparative analysis of visual and numerical outcomes among DLG, GGI, ROG, and $\textbf{AFGI}$. To address the mismatch between $\hat{y}$ and $x$ in large batch sizes, we sort $\hat{y}$ in an ascending order, following  settings the same as those in GradInversion~\cite{yin2021see}. The $MI$ for each strategy is set according to their respective previous studies. Visual representations of all $\hat{x}$ are presented in Fig.~\ref{fig:fig 3}. To quantitatively assess performance, we employ three evaluation metrics: PSNR, SSIM, and LPIPS, which are summarized in Table~\ref{tab:tab3}. 

From Fig.~\ref{fig:fig 3}, the findings consistently demonstrate that $\textbf{AFGI}$ outperforms other strategies in terms of SSIM and LPIPS. Image objects can be accurately identified in Fig.s~\ref{fig:fig 3} (b) and (g). In contrast, DLG~\cite{zhu2019deep} (as shown in Fig.~\ref{fig:fig 3} (c) and (h)) and ROG (as shown in Fig.~\ref{fig:fig 3} (e) and (j)) fail to detect objects, while the color in GGI (as shown in Fig.~\ref{fig:fig 3} (d) and (i)) is distorted. Moreover, $\textbf{AFGI}$ is of a  faster model convergence in reconstructing useful image information (such as image objects) compared to all other strategies. Despite consuming less time in ResNet-50, ROG~\cite{yue2023gradient} results are worse in identifying image subjects. These outcomes align with previous findings from GGI and GradInversion~\cite{geiping2020inverting, yin2021see}, which suggest that $\hat{x}$ in shallow models is better than that in deep models.

In Table~\ref{tab:tab3}, it is evident that  ROG fails to detect any objects or useful information in the images (as shown in Fig.s~\ref{fig:fig 3} (e) and (j)), though its PSNR values remain higher than those of other methods. This result indicates that smoothing images is helpful for improving PSNR values. This also suggests that  solely relying on traditional evaluation metrics, such as PSNR, does not precisely gauge the true quality of images. $\textbf{AFGI}$ achieves better SSIM and LPIPS values in all $\hat{x}$. We conjecture that the use of the encoder and decoder to smooth $\hat{x}$ makes the image edges to blurred  and immersed with the image background, resulting in low SSIM and LPIPS values.

Finally, we present additional reconstructed results for the indexes of 5000, 7000, and 9000 images in Fig.~\ref{fig:1000th} to illustrate the universality of the $\textbf{AFGI}$ strategy.

\begin{table}[b]
\caption{The metric values of Fig.~\ref{fig:ablation}}
\centering
\renewcommand{\arraystretch}{1.1}
\begin{tabular}{ccccc}
\cline{1-4}
\multirow{2}{*}{\textbf{$\mathbf{R}_{reg}$} }                 & \multicolumn{3}{c}{\textbf{Metrics}}                                                                                                 \\ \cline{2-4} 
                & \multicolumn{1}{c}{\textbf{PSNR $\uparrow$}} & \multicolumn{1}{c}{\textbf{SSIM  $\uparrow$}} & \multicolumn{1}{c}{\textbf{LPIPS $\downarrow$ }} \\\cline{1-4} 
\multirow{2}{*}{{\textbf{None}}} &14.51                     &        0.0092         &            0.7576                 \\
                           &14.94                    &      0.0091               &        0.7039                      \\ \hline
                           
\multirow{2}{*}{{\textbf{$+\mathbf{R}_{TV}$}} }& 16.08                     &       0.0591              &         0.6068               \\
                           
                           &      \textbf{16.32}           &       0.0820         &       0.5899                 \\ \hline
\multirow{2}{*}{{\textbf{$+\mathbf{R}_{mean}$}} }
                           &16.30                   &               \textbf{0.0514 }       &          0.5640                 \\
                           & 16.07                     &            0.1065       &   0.6116              \\ \hline
\multirow{2}{*}{{\textbf{$+\mathbf{R}_{ed}$}} }
                           & \textbf{17.47}                   &  0.0572                     & \textbf{0.4909}                            \\
                           & 15.87                    & \textbf{0.1183}                     & \textbf{0.5762}                       \\ \hline
\end{tabular}
\label{tab:tab4}
\end{table}

\begin{figure}[t]
\centering
 \subfloat[\centering  Ground-truth]{
    \includegraphics[width=0.085\textwidth]{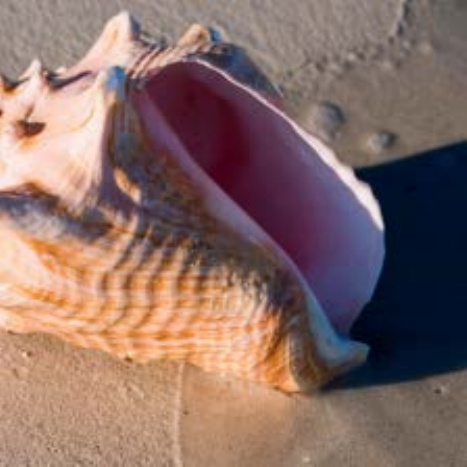}
    \label{fig:ablation-a}
}
\subfloat[\centering  None]{
    \includegraphics[width=0.085\textwidth]{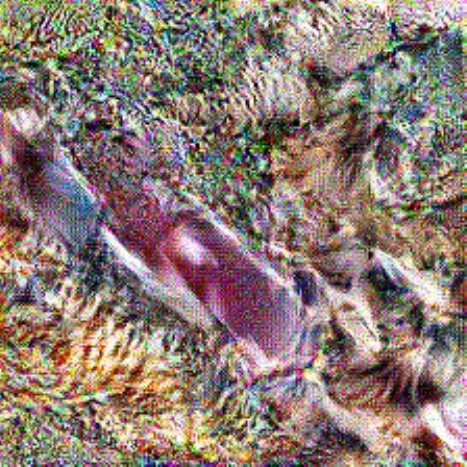}
    \label{fig:ablation-b}
}
\subfloat[\centering  +$\mathbf{R}_{TV}$]{
    \includegraphics[width=0.085\textwidth]{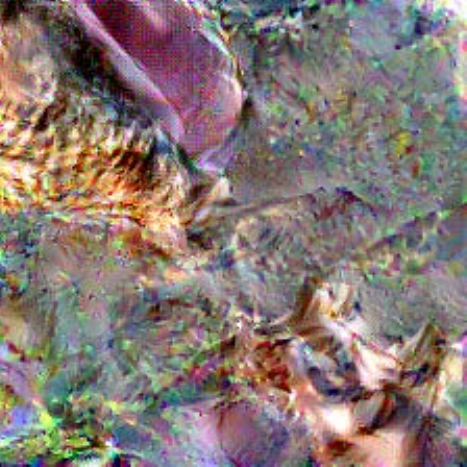}
    \label{fig:ablation-c}
}
\subfloat[\centering  +$\mathbf{R}_{mean}$]{
    \includegraphics[width=0.085\textwidth]{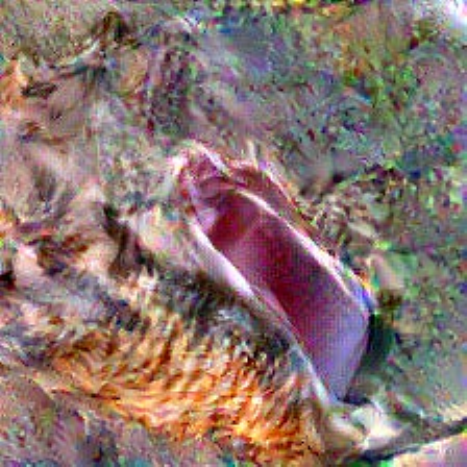}
    \label{fig:ablation-d}
}
\subfloat[\centering  +$\mathbf{R}_{ed}$]{
    \includegraphics[width=0.085\textwidth]{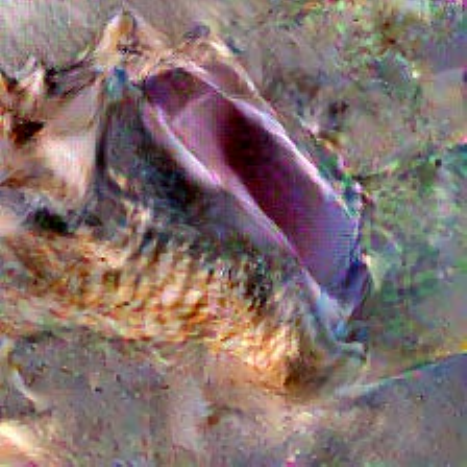}
    \label{fig:ablation-e}
}
  \\
  \subfloat[\centering  Ground-truth]{
    \includegraphics[width=0.085\textwidth]{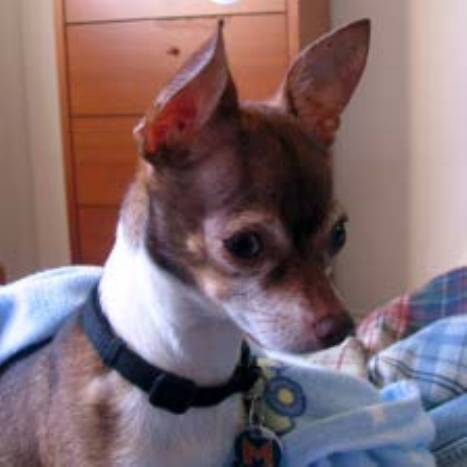}
    \label{fig:ablation-a}
}
  \subfloat[\centering  None]{
    \includegraphics[width=0.085\textwidth]{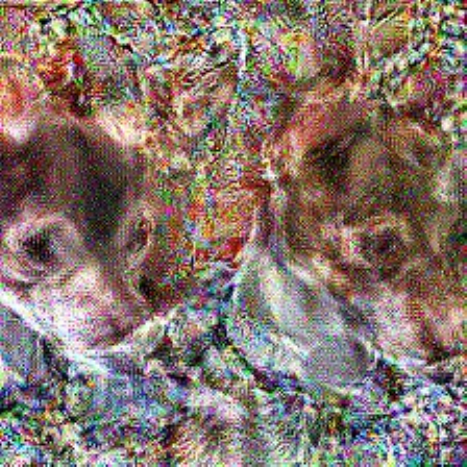}
    \label{fig:ablation-b}
}
\subfloat[\centering  +$\mathbf{R}_{TV}$]{
    \includegraphics[width=0.085\textwidth]{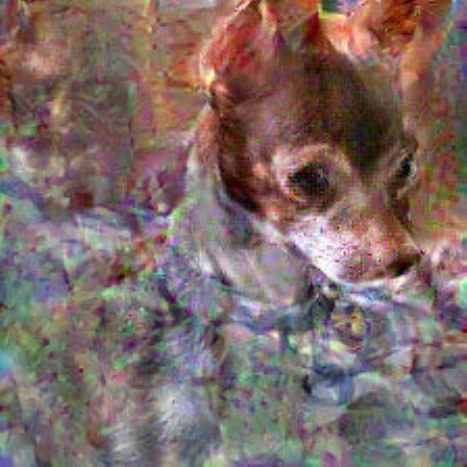}
    \label{fig:ablation-c}
}
\subfloat[\centering  +$\mathbf{R}_{mean}$]{
    \includegraphics[width=0.085\textwidth]{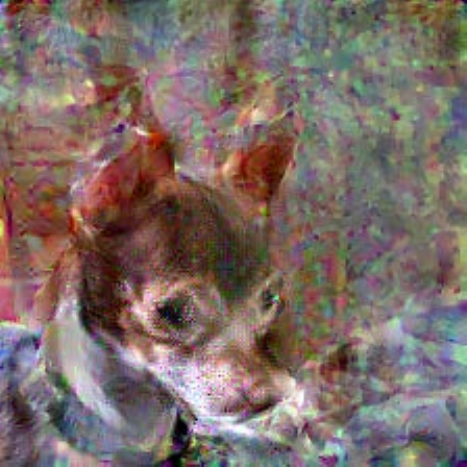}
    \label{fig:ablation-d}
}
\subfloat[\centering  +$\mathbf{R}_{ed}$]{
    \includegraphics[width=0.085\textwidth]{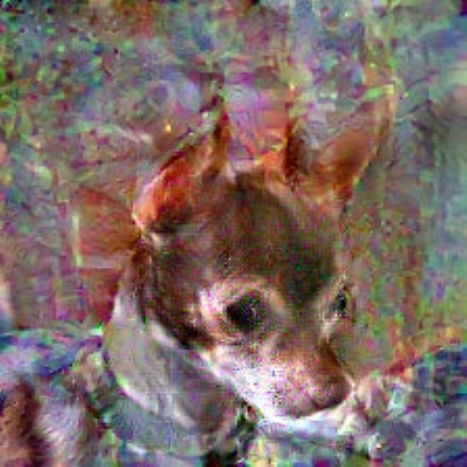}
    \label{fig:ablation-e}
}
\caption{Ablation study of each regularization term at a batch size of 1. Ground-truth images are shown in figure (a) and (f), while the others are reconstructed using different regularization terms.}
\label{fig:ablation}
\end{figure}

\textbf{Ablation experiment:}
To elucidate the effect of each regularization term on $\hat{x}$, we sequentially incorporate regularization term one by one into the reconstruction process. A visual and quantitative comparison of $\hat{x}$ is summarized in Fig.~\ref{fig:ablation} and Table~\ref{tab:tab4}. If there is no regularization term in GIAs, it is denoted as None in Fig.~\ref{fig:ablation} (b).  The $\hat{x}$ results exhibit significant pixel noises. The regularization term $\mathbf{R}_{TV}$ enhances image quality and prohibits noises. However, $\mathbf{R}_{TV}$ introduces a noticeable shift of the subject's position and inaccurate colors in $\hat{x}$, as shown in Fig.~\ref{fig:ablation} (c). In  Fig.~\ref{fig:ablation} (d), we introduce $\mathbf{R}_{mean}$ to correct color in $\hat x$. This operation can increase the PSNR values of $\hat x$.  Finally, to rectify the position of the image subject, the objective function incorporates $\mathbf{R}_{ed}$.  The results show higher SSIM values and lower LPIPS values in most cases in Table~\ref{tab:tab4}, sheding light on the overall improvement of recovered $\hat{x}$.

\begin{figure}[t]
  \centering

 \subfloat[\centering  Ground-truth]{
    \includegraphics[width=0.11\textwidth]{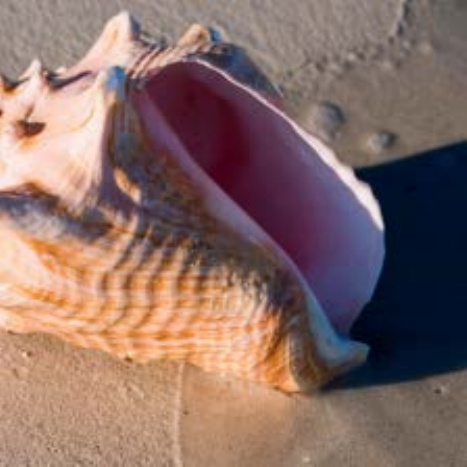}
    \label{fig:gray-a}
}
\subfloat[\centering  $\textbf{AFGI}$]{
    \includegraphics[width=0.11\textwidth]{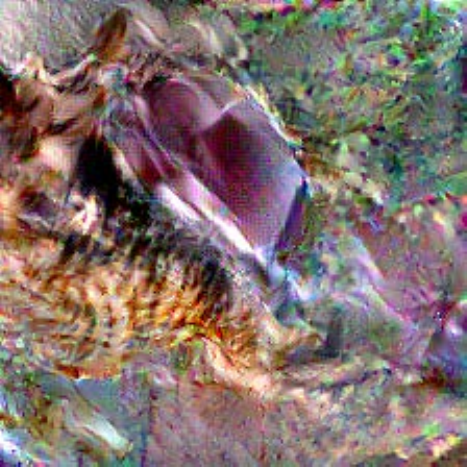}
    \label{fig:gray-b}
}
\subfloat[\centering  $\textbf{AFGI}$-r]{
    \includegraphics[width=0.11\textwidth]{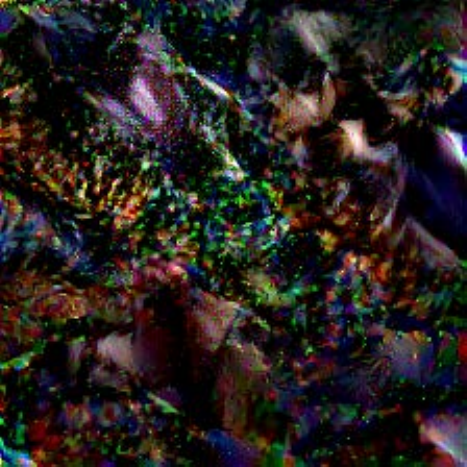}
    \label{fig:gray-c}
}
\subfloat[\centering  $\textbf{AFGI}$-$\ell_2$]{
    \includegraphics[width=0.11\textwidth]{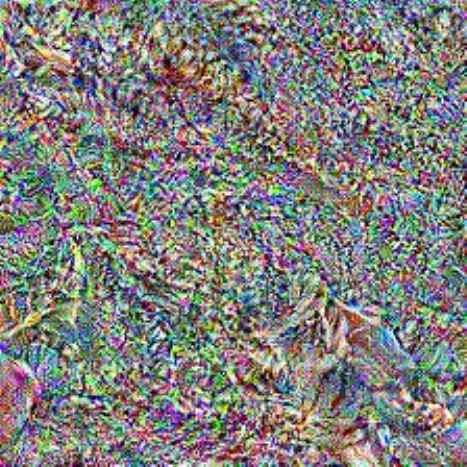}
    \label{fig:cost-c}
}
 
\subfloat[\centering Ground-truth]{
    \includegraphics[width=0.11\textwidth]{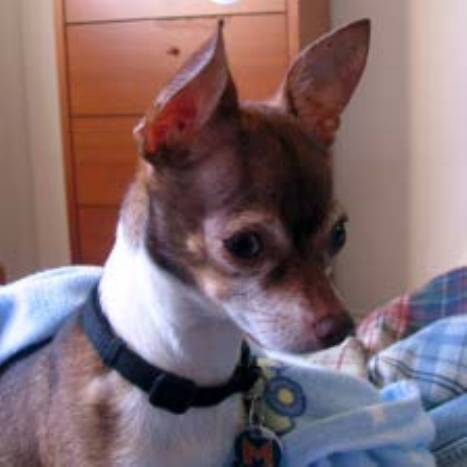}
    \label{fig:gray-h}}
\subfloat[\centering  $\textbf{AFGI}$]{
    \includegraphics[width=0.11\textwidth]{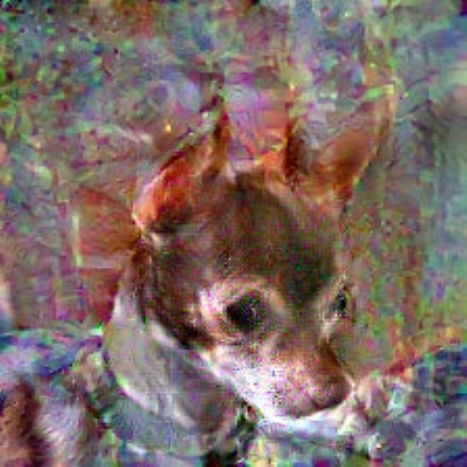}
    \label{fig:gray-i}
}
\subfloat[\centering  $\textbf{AFGI}$-r]{
    \includegraphics[width=0.11\textwidth]{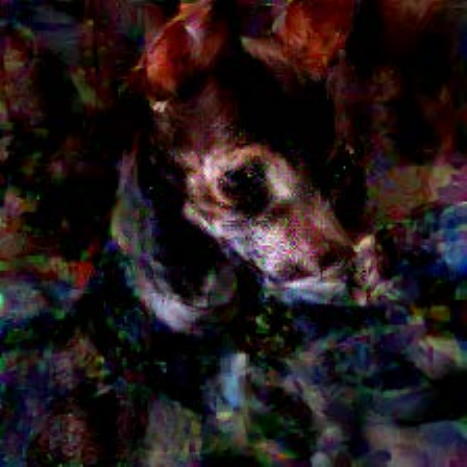}
    \label{fig:gray-j}
}
\subfloat[\centering  $\textbf{AFGI}$-$\ell_2$]{
    \includegraphics[width=0.11\textwidth]{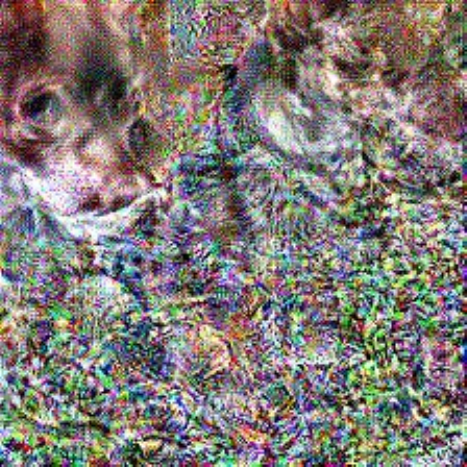}
    \label{fig:cost-j}
}
  \caption{The $\hat x$ of $\textbf{AFGI}$ based on different initializations of $x$ and different loss functions.}
  \label{fig:gray}
\end{figure}

\textbf{Impact of image initialization:} 
To examine the impact of $\hat x$ initialization on  final $\hat x$, visual results and evaluation metrics are presented in Fig.~\ref{fig:gray} (b) (e) and Table~\ref{lab:gray}. A notable finding is the superior reconstruction achieved when using gray images as the initial value of  $\hat{x}$. Comparisons between random initialization of $\hat{x}$ ($\textbf{AFGI}$-r) and gray image initialization of $\hat{x}$ ($\textbf{AFGI}$) highlight the superior reconstruction achieved when using gray images as the initial $\hat{x}$. This underscores the vital role of initial $\hat{x}$ in enhancing the quality of final reconstruction.

\begin{table}[h]
\caption{The metric values for Fig.~\ref{fig:gray}}
\centering
\renewcommand{\arraystretch}{1.2}
\begin{tabular}{ccccc}
\hline
\multirow{2}{*}{{\textbf{$\hat x$}}     }      & \multicolumn{3}{c}{\textbf{Metrics}}                                 \\ \cline{2-4} 
                                     & \textbf{PSNR $\uparrow$}  & \textbf{SSIM $\uparrow$}   & \textbf{LPIPS $\downarrow$}    \\ \cline{1-4} 
\multirow{2}{*}{{\textbf{AFGI}}} & \textbf{17.47} & \textbf{0.0572} & \textbf{0.4909}\\
                                     &  \textbf{15.87}          &  \textbf{0.1183 }         & \textbf{0.5762}  \\ \hline
\multirow{2}{*}{\textbf{AFGI}-r }& 12.54          & 0.0171          & 0.6656              \\
                                     & 10.33          & 0.0393          & 0.5922          \\ \hline
\multirow{2}{*}{\textbf{AFGI}-$\ell_2$ }& 14.00          & 0.0089          & 1.1852           \\
                                          & 14.29          & 0.0081          & 0.9543              \\ \hline
\end{tabular}
 \label{lab:gray}
\end{table}

\textbf{Impact of loss functions:}
We conduct fine-tuning experiments, as shown in Fig.~\ref{fig:gray}, to examine the impact of different cost functions. Our experimental results indicate that the cosine similarity cost function achieves the highest  quality of $\hat{x}$, providing a notable advantage. We consider $\textbf{AFGI}$-$\ell_2$ by modifying the cost function to cosine similarity and $\ell_2$ distance. Visual results of reconstructed images and corresponding metrics are provided in Fig.~\ref{fig:gray} (c) and (f), and the last row of Table~\ref{lab:gray}. Notably, cosine similarity exhibits a stronger tendency to generate high-quality $\hat{x}$. All strategies based on the cosine similarity cost function show higher PSNR values and lower LPIPS values than those using the $\ell_2$ distance cost function. This observation deviates from the findings in GradInversion~\cite{yin2021see}. The discrepancy is attributed to different regularization terms. In our experiments, the choice of the cost function can significantly influence the quality of recovered  $\hat{x}$, with cosine similarity showing a substantial advantage.

\begin{figure}[t]
\centering
\includegraphics[width=0.8\columnwidth]{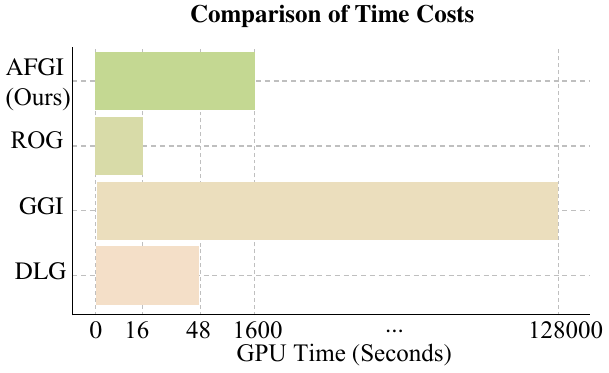}
\caption{The time costs of different strategies at the batch size =1. }
\label{fig:time}
\end{figure}

\textbf{Time cost comparison:}
Finally, we compare the time costs of DLG~\cite{zhu2019deep}, GGI~\cite{geiping2020inverting}, ROG~\cite{yue2023gradient}, and $\textbf{AFGI}$ when the batch size is 1. The results are shown in Fig.~\ref{fig:time}. Notably, $\textbf{AFGI}$ consistently outperforms GGI in terms of reconstruction quality, even with a reduction of over 85\% in total time costs. Although ROG and DLG have lower costs, they perform poorly and fail to extract useful information from $\hat{x}$. These two methods are considered ineffective in reconstructing high-resolution images in deep models. $\textbf{AFGI}$ strikes a balance between time costs and better GIA final results for $\hat{x}$.

\begin{figure}[t]
  \centering
 \subfloat[\centering  Ground-truth]{
    \includegraphics[width=0.45\textwidth]{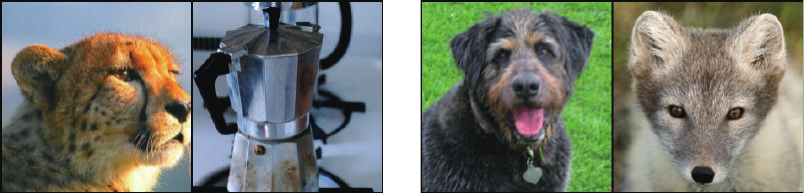}
    \label{fig:BS=2-a}
}

\subfloat[\centering  \textbf{AFGI}]{
    \includegraphics[width=0.45\textwidth]{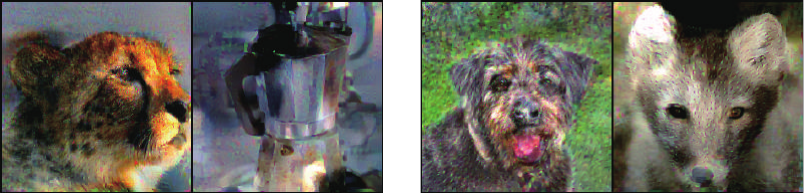}
    \label{fig:BS=2-b}
}
  \caption{The $\hat x$ of  $\textbf{AFGI}$ under a batch size is 2. }
  \label{fig:BS=2}
\end{figure}

\subsubsection{Reconstructed Results as Batch Size $>$ 1}
Fig.~\ref{fig:BS=2} displays the $\hat{x}$ results of $\textbf{AFGI}$ with a training batch size of 2. It is easy to observe that the color of $\hat{x}$ in the third column of Fig.~\ref{fig:BS=2} (b) is lighter than the corresponding ground-truth data, while the color of the image in the fourth column is darker than its ground-truth counterpart. It is important to note that $\hat{x}$ is influenced by factors such as the color and image subjects of other images in the same training batch.

\begin{figure}[t]
  \centering
\subfloat[\centering  Ground-truth]{
    \includegraphics[width=0.085\textwidth]{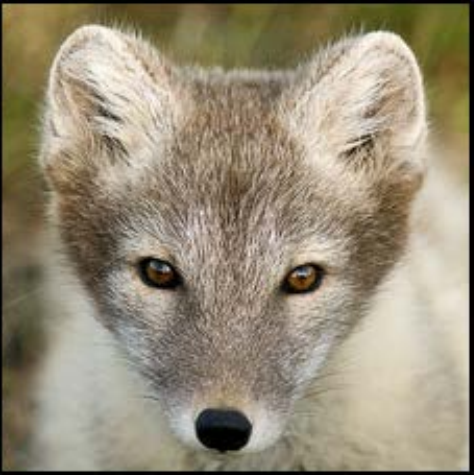}
    \label{fig:BS=48-a}
}
 \subfloat[\centering  batch size = 1]{
    \includegraphics[width=0.085\textwidth]{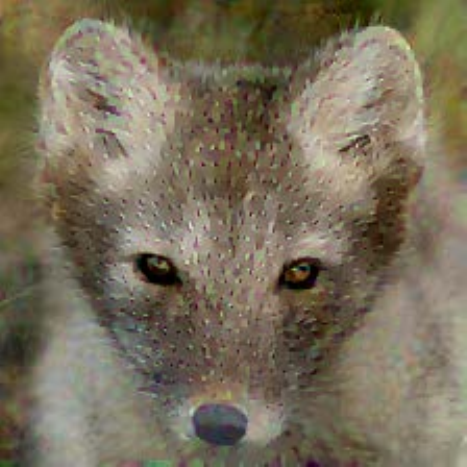}
    \label{fig:BS=48-b}
}
\subfloat[\centering  batch size = 2]{
    \includegraphics[width=0.085\textwidth]{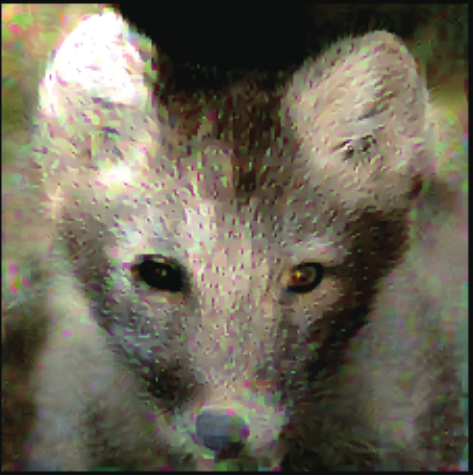}
    \label{fig:BS=48-c}
}
\subfloat[\centering  batch size = 8]{
    \includegraphics[width=0.085\textwidth]{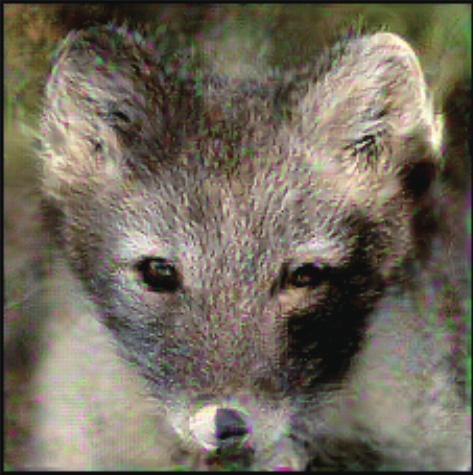}
    \label{fig:BS=48-d}
}
\subfloat[\centering  batch size = 48]{
    \includegraphics[width=0.085\textwidth]{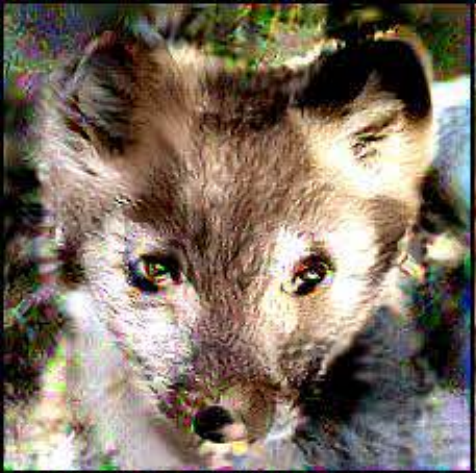}
    \label{fig:BS=48-e}
}
  \caption{The $\hat x$ results of $\textbf{AFGI}$ with batch sizes ranging from 1 to 48.}
  \label{fig:BS=48}
\end{figure}

Next, we evaluate the attack performance of $\textbf{AFGI}$ with the batch size varying  from 1 to 48. The results are presented in Fig.~\ref{fig:BS=48}. $\textbf{AFGI}$ maintains its ability to visually identify the image subject in $\hat{x}$ even with a large training batch size. However, the color of $\hat x$ in a large training batch may shift due to the influence of other images in the same batch. From these results,  it is evident  that the quality of $\hat{x}$ deteriorates as the batch size increases. This finding also implies that increasing the batch size can better preserve data privacy.

\section{Conclusion}
\label{CON}

We introduce $\textbf{AFGI}$, an effective iteration-based gradient inversion strategy that combines two novel approaches: $\textbf{LRB}$ to enhance label accuracy and $\textbf{VME}$ to improve the quality of reconstructed images. $\textbf{LRB}$ utilizes a convolutional network to accurately reconstruct labels without the strong assumption of no-repeating labels in a training batch. Experimental results highlight the efficiency of $\textbf{AFGI}$ in significantly reducing time costs and consistently achieving superior reconstructed results. These findings manifest the practicality of gradient inversion in real-world applications and underscore the necessity to safeguard gradients.

In our  future work, we will extend our algorithm into the domain of text reconstruction. Besides, we will also explore defense strategies to counteract these gradient inversion attacks so as to boost user privacy preservation. 

\bibliographystyle{IEEEtran}
\bibliography{ref}

\vfill

\end{document}